\documentclass[lettersize,journal]{IEEEtran}
\usepackage{amsmath,amsfonts}
\usepackage{algorithmic}
\usepackage{algorithm}
\usepackage{array}
\usepackage[caption=false,font=normalsize,labelfont=sf,textfont=sf]{subfig}
\usepackage{textcomp}
\usepackage{stfloats}
\usepackage{url}
\usepackage{verbatim}
\usepackage{graphicx}
\usepackage{cite}
\usepackage{soul}
\usepackage{color}
\usepackage{ulem}
\usepackage{here}
\usepackage{empheq}
\usepackage[super]{nth}

\begin{document}

\title{Active Inference in Contextual Multi-Armed Bandits for Autonomous Robotic Exploration}

\author{Shohei Wakayama$^1$, Alberto Candela$^2$, Paul Hayne$^3$, and Nisar Ahmed$^1$
\thanks{$^1$S. Wakayama and N. Ahmed are with the Smead Aerospace Engineering Sciences Department, University of Colorado Boulder, Boulder, CO 80303 USA        {\tt\small [shohei.wakayama; nisar.ahmed]@colorado.edu}}
\thanks{$^2$A. Candela is with the Jet Propulsion Laboratory, California Institute of Technology, Pasadena, CA 91109, USA {\tt\small alberto.candela.garza@jpl.nasa.gov}}
\thanks{$^3$P. Hayne is with the Astrophysical \& Planetary Sciences Department, University of Colorado Boulder, CO 80303 USA {\tt\small paul.hayne@lasp.colorado.edu}}
}

\markboth{IEEE Trans. Robotics,~Vol.~V, No.~N, November~2024}%
{Wakayama and Ahmed: Active Inference Contextual Multi-Armed Bandits}


\maketitle

\newcommand{\sw}[1]{\textcolor{red}{[sw: #1]}}
\newcommand{\na}[1]{\textcolor{blue}{[na: #1]}}

\newcommand{\edt}[1]{\textcolor{magenta}{#1}}

\begin{abstract}
Autonomous selection of optimal options for data collection from multiple alternatives is challenging in uncertain environments. When secondary information about options is accessible, such problems can be framed as contextual multi-armed bandits (CMABs). Neuro-inspired active inference has gained interest for its ability to balance exploration and exploitation using the expected free energy objective function. Unlike previous studies that showed the effectiveness of active inference based strategy for CMABs using synthetic data, this study aims to apply active inference to realistic scenarios, using a simulated mineralogical survey site selection problem. Hyperspectral data from AVIRIS-NG at Cuprite, Nevada, serves as contextual information for predicting outcome probabilities, while geologists' mineral labels represent outcomes. Monte Carlo simulations assess the robustness of active inference against changing expert preferences. Results show active inference requires fewer iterations than standard bandit approaches with real-world noisy and biased data, and performs better when outcome preferences vary online by adapting the selection strategy to align with expert shifts.

\end{abstract}


\section{Introduction} \label{sec: introduction}
For robotic exploration of uncertain environments such as outer solar system planets, disaster sites, and geologically intriguing areas on Earth, it is often crucial to optimally select among multiple alternative options to enable autonomous data collection (e.g. selecting a mineral rock specimen for detailed chemical analysis and landing site selection for a planetary rover) \cite{grant2018science}. Such decision making has been mostly performed by human domain experts, such as scientists and engineers \cite{practicingMars2020}, since this mitigates the possible dangers posed to exploration robots which are costly and difficult to replace. However, this approach leads to significant mental workload on humans \cite{europaclippercost2019, mars2020project}. Additionally, due to the difficulty of interpreting low-quality data sent from the robots, there is a risk that humans might overlook optimal options and make suboptimal decisions, ultimately decreasing mission efficiency. Moreover, for highly remote and underexplored uncertain environments (e.g. icy moons of Jupiter and disaster sites at a nuclear power plant), it is not possible to rely on frequent and information-rich human-robot interaction because of the significant distance between humans and robots, limited bandwidth in communication, and necessity for increased ``housekeeping" downtime for the robots. Therefore, robots operating in uncertain environments are expected to efficiently and autonomously determine the best options for data collection to mitigate the aforementioned issues and to enhance the overall mission outcomes. Nevertheless, it is not straightforward to make such decisions due to the stochastic nature of the dispatched environments, since sensing outcomes are stochastic and the distributions of the outcome observations are unknown {\it a priori}. 


\begin{figure}[t]
    \centering
    \includegraphics[width=0.9\linewidth]{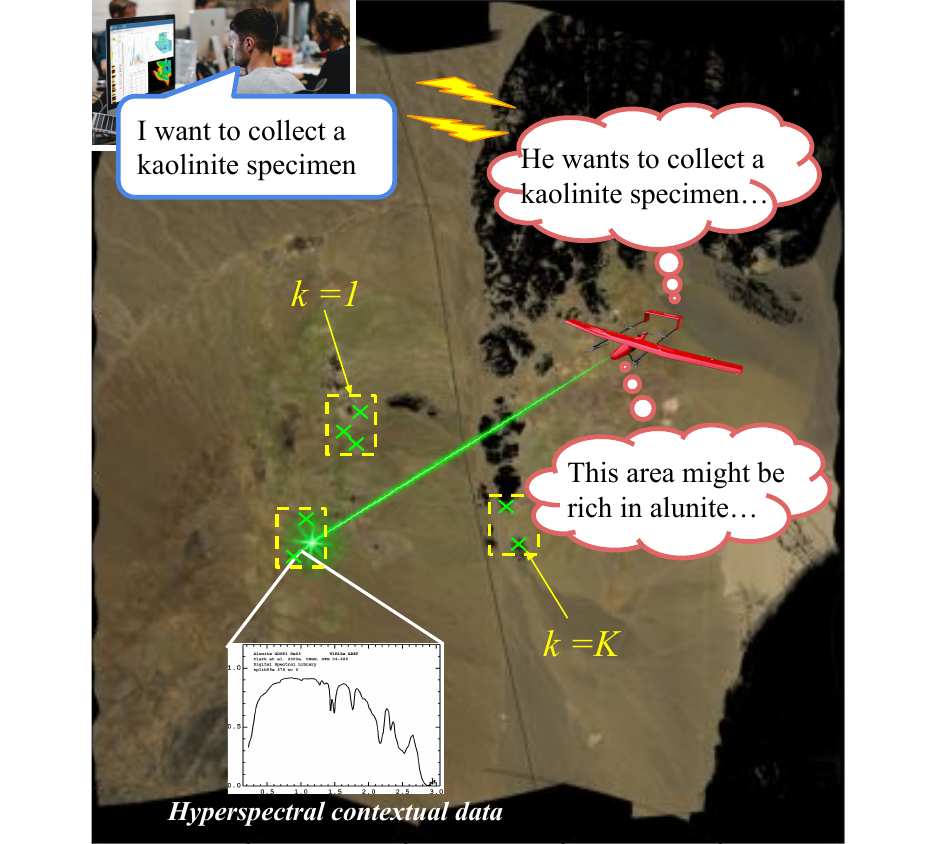}
    \captionsetup{skip=5pt}
    \caption{An aerial robot reasons where a desired mineral rock specimen can be sampled by utilizing remote sensing data such as hyperspectral information. To speed up the process, it is desirable not only to strike a balance between exploitation and exploration, but also incorporate a selection bias, i.e. human expert prior preferences, regarding the desired observations.}
    \label{fig: scenario}
\end{figure}

For instance, consider a scenario, as illustrated in Fig. 1, where an aerial robot must autonomously select the most promising site for sampling a desired mineral rock specimen during a follow-up in-situ survey mission \cite{muirhead2020mars}. This search site selection is based on remote sensing data (e.g. hyperspectral information) gathered from predetermined search sites using a lightweight sensor (e.g. spectrometer). In this scenario, however, the sensing returns obtained by directing sensors are stochastic for various reasons, for example, due to observation noise and the variability in the targeted coordinates for each site sampling instance. Moreover, model parameters used to predict the likelihood of observing each outcome (e.g. a detected mineral specimen) based on these returns are not known beforehand. Thus, the robot needs to carefully strike a balance between {\it exploitation} (increasing the certainty in sites where desired mineral specimens are expected) and {\it exploration} (decreasing the uncertainty in poorly explored search sites). 
This sequential decision-making problem can be formulated with mathematical frameworks such as partially observable Markov decision processes (POMDPs) \cite{kaelbling1998, kurniawati2022} and Gaussian processes (GPs) \cite{krause2008near, candela2021using}. In the case of POMDPs, however, careful definitions of a state transition function, a reward function, and planner hyperparameters (e.g. a planning depth and a discount factor) are required. Reward and hyperparameter tuning, in particular, can be tricky and time-consuming in new and uncertain environments \cite{denniston2023}. On the other hand, in the case of GPs, while they can leverage a spatial structure of an environment, the computation cost of kernel functions is significant for real-time operation \cite{rasmussen2003gaussian, west2021use}. Additionally, neither framework easily incorporates human expert preferences regarding outcomes easily. Thus, instead, we opt to study simpler contextual multi-armed bandit (CMAB) formulation, which has been widely studied in recommendation systems, finance, healthcare, and recently, robotics \cite{li2010, zhou2015survey, bouneffouf2020, rudra2023contextual}, and allows us to advantageously abstract certain lower-level behavioral aspects of the search site selection problem.



However, in general, bandit problems--depending on their scale and complexity--often require a large number of iterative interactions with the environment to finalize the optimal option. This need for numerous iterations can become a bottleneck when applying this mathematical framework to practical robotic applications, such as space exploration and mineralogical surveys on Earth, where resources and time are often constrained. Moreover, existing conventional methods in CMABs typically do not explicitly take into account human experts' (e.g. scientists') prior preferences regarding outcomes in their decision-making processes. As a consequence, the decisions derived from these methods may not always align with what humans are actually interested in observing, leading to the decrease in mission efficiency. In light of these backdrops, our previous studies \cite{wakayamaICRA2023, wakayama2023observationaugmented} sought to emulate the approach taken by astronaut Harrison Schmitt during the Apollo 17 mission, where he combined in-situ findings with geological expert knowledge to advance lunar geology \cite{schmitt1973}. To achieve a similar behavior in robotic systems, we applied active inference (AIF) \cite{smith2022step, parr2022, lanillos2021active}--which originated in computational neuroscience and has recently gained traction in robotics--to develop option selection strategies for stationary, independent, and linear CMABs that are informed by expert-provided prior preferences on observations. While these studies showed that AIF agents could efficiently identify the best option for humans compared to other strategies when expert prior preference is {\it stationary}, the contextual information used for decision-making and the true hidden model parameters associated with the options were randomly generated, which does not reflect real-world conditions. Hence, in this article, we aim to validate the applicability of the AIF-based option selection methodology in realistic problem scenarios. Specifically, the key contributions and novelty with respect to previous studies \cite{wakayamaICRA2023, wakayama2023observationaugmented} are: 

\begin{itemize}
    \item Demonstrating the effectiveness of the AIF-based option selection algorithm using real scientific data, namely based on hyperspectral data collected by AVIRIS-NG \cite{hamlin2011} and mineral label data created by geologists \cite{swayze2014mapping}. This study marks the first application of active inference in geological data exploration.
    \item Showcasing the superiority of the proposed method to conventional bandit option selection strategies even when human expert's preferences for desired outcomes change {\it dynamically}.
\end{itemize}

The remainder of this paper is organized as follows. Sec. \ref{sec: background} provides an overview of multi-armed bandits (MABs) and CMABs, along with an introduction to active inference. Sec. \ref{sec: methodology} describes the problem statement, and then presents the AIF-based option selection method for CMABs. In Sec. \ref{sec: simulation_study}, we explain the science dataset and detail the preprocessing procedures. Following that, we present the offline training results used to learn the ``true" (but unknown to the exploration robot) hidden parameters associated with options. Then, the simulation setup is outlined and the results from the simulated Monte Carlo experiments are discussed. Finally, Sec. \ref{sec: conclusion} concludes the study with a summary of key findings and the potential research directions.

\section{Background} \label{sec: background}
\subsection{Multi-Armed Bandits (MABs) and Contextual MABs} \label{sec: cmabs} 
The multi-armed bandit (MAB) is a classic reinforcement learning problem that involves identifying and utilizing the optimal option among multiple alternatives \cite{mahajan2008multi}. In MABs, an outcome from each option (a.k.a. ``arm") is probabilistic, and its distribution is unknown {\it a priori}, leading to the so-called exploration-exploitation dilemma since only one option's outcome can be observed per decision-making iteration. Therefore, bandit agents repeatedly execute two key steps--1) option/arm selection and 2) measurement update--to ideally minimize the cumulative regret, which measures the disparity between the total reward achieved by consistently selecting the best option (unknown during execution) and that obtained following a specific option selection strategy \cite{kuleshov2014}. In standard MAB, however, since the information used for option selection is {\it solely} based on past outcome observations, a sufficiently large number of iterations is typically required to identify the best option. 

Conversely, in contextual MABs (CMABs), additional side information, known as {\it contexts}, associated with each option is used to predict outcome observation probabilities. This prediction is done in conjunction with the unique hidden parameters of each option during option selection, and allows for more efficient identification of the optimal option and minimization of the cumulative regret. For measurement updates, Bayes' theorem is primarily used. For option selection, $\varepsilon$-greedy, strategies based on the upper confidence bound (UCB) \cite{auer2002, kaufmann2018bayesian}, Thompson sampling \cite{thompson1933, agrawal2013thompson}, and methods using the softmax function \cite{kuleshov2014} are well-known. However, these conventional option selection methods often rely on heuristics to achieve good performance. Additionally, external preferences, such as those from domain experts regarding valuable outcome observations for robotic exploration, cannot be easily incorporated. Hence, the outcomes obtained by following these option selection strategies may not align with what humans actually want to observe. Therefore, it is important to develop an alternative option selection strategy particularly for such robotics applications. This is because the number of iterations must generally be limited in such applications, and incorporating human prior preferences in decision-making is crucial to enhance the interpretability of the robot's decisions.

\subsection{Free Energy Principle and Active Inference} \label{sec: fep_aif} 
The free energy principle (FEP) is a theoretical framework proposed in the field of computational neuroscience to mathematically and systematically explain the functioning of the brain \cite{friston2006free, friston2010free}. According to this principle, biological agents form probabilistic internal models of external environments and, based on these models, perceive, learn, and act to minimize the discrepancy (i.e. free energy) between predicted observations and actual sensory inputs, thereby increasing their chances of survival. Predictive coding, known in research on the visual cortex, is one specific implementation of the FEP \cite{friston2009predictive, huang2011predictive}.  

Active inference (AIF) is a mathematical framework that applies the FEP specifically to the behavioral norms of biological agents \cite{smith2022step, parr2022}. In the field of neuroscience, it has been used to understand the characteristic behavioral mechanisms observed in patients with autism \cite{arthur2021examination, arthur2023testing}. Recently, it has also garnered attention in the field of robotics for state estimation, adaptive control, and for decision making under uncertainty \cite{pio2016active, pezzato2020, lanillos2021active, baioumy2021}. The reason for this lies in the {\it expected free energy (EFE)} objective function characterizing AIF. Although a detailed explanation is provided in Sec. \ref{sec: methodology}, the EFE for each possible option/action in the MAB/CMAB context consists of the (negative) value resulting from selecting a particular option and the (negative) information gain (i.e. {\it mutual information} commonly known in robotics \cite{julian2014mutual, maani2015} and Bayesian experimental design \cite{chaloner1995bayesian, huan2013simulation}) representing how much the uncertainty about hidden states is reduced by taking that option. Consequently, by optimizing (i.e. minimizing) the EFE, agents can naturally take an action balancing exploitation and exploration. Additionally, since preference information regarding outcome observations, known as {\it prior preference}, can be externally incorporated into the value term, agents take actions biased towards obtaining desired observations. This characteristic has recently been studied for Pareto point selection problems in multi-objective reinforcement learning\cite{amorese2024online}.

Given these backdrops, our previous works have proposed AIF-based option selection methods for CMABs, particularly when hybrid discrete-continuous observation likelihoods such as sigmoid and softmax functions are employed \cite{wakayamaICRA2023, wakayama2023observationaugmented}. Although autonomous robotic agents with these methods occasionally get stuck in local minima due to selection bias, extensive simulation experiments with synthetic datasets have demonstrated that the AIF agents can identify the best option with a far fewer number of iterations. Yet, the practicability of the proposed AIF methods has not been validated on more realistic data. Also, in our previous studies, the prior preference distributions are assumed to be stationary. However, in realistic scenarios, human preferences regarding outcomes can change dynamically. Therefore, after introducing the CMAB problem and reviewing the proposed AIF-based option selection method, we are going to present how the method is validated and demonstrated for more practical problems with these characteristics, using a real scientific dataset. 




\section{Methodology} \label{sec: methodology}

\subsection{Problem Statement} \label{sec: problem_statement} Suppose the total number of options (e.g. search sites) taken into account by a robot is $K \in \mathbb{N}$. Note that these options are equivalent to the bandit arms and selecting an option $k \in \{1, \cdots, K\}$ is denoted as $a = k$ (for the ease of notation, in the following, we use $a_k \leftrightarrow a=k$). Additionally, suppose that a semantic observation $o_k$ (e.g. mineral label) of each option $k$ from an observation source is multicategorical across $F$ labels, i.e. $o_k = f, f\in \mathcal{F} = \{1, \cdots, F\}$. Therefore, the probability that a feature $f$ is observed by investigating an option $k$ at a decision instance $t$ can be described as the following softmax likelihood function \cite{bishop2006, nisar2018}\footnote{It is also natural to choose a Dirichlet distribution as a prior and a categorical distribution as an observation likelihood, as their conjugacy allows for easy posterior calculation \cite{bishop2006}. Nevertheless, this approach cannot easily incorporate continuous contextual information (such as hyperspectral data) associated with the options. Hence, the softmax function, which is one of hybrid discrete-continuous likelihood functions and has gained attention in the field of multi-sensor fusion \cite{nisarTRO, sweet2016, tse2018, lukeTRO}, is adopted.}, 
\begin{align} \label{eq: softmax}
    p(o_{k,t}=f|\vec{\Theta}_k; \vec{x}_{k,t}) = \frac{e^{\vec{w}_{k,f}^T \vec{x}_{k,t} + b_{k,f}}}{\sum_{h=1}^{F} e^{\vec{w}^T_{k,h} \vec{x}_{k,t} + b_{k,h}}},
\end{align}
where $\vec{\Theta}_k = [\vec{w}_{k,1}, b_{k,1}, \cdots, \vec{w}_{k,F}, b_{k,F}], \vec{\Theta}_k \in \mathbb{R}^{(C+1)\times F}$ is a hidden linear parameter vector unique to the option $k$, and $\vec{x}_{k,t} \in \mathbb{R}^{C}$ is a (dynamic) context vector (e.g. indicating the choice of in-situ hyperspectral measurement) specific to the option $k$, where $C$ is the context feature dimension (e.g. the number of available hyperspectral bands).  

Recall that the objective of CMABs is to minimize cumulative regret. Here, a unit reward $(1)$ is provided if a predetermined preferable feature $f_p \in \mathcal{F}$ is observed for $o_k$, and no reward $(0)$ is given if any other feature is observed. In the case of the search site selection scenario, for instance, $f_p$ represents a particular mineral label that scientists want the robot to investigate (e.g. a kaolinite specimen). Thus, if the probability of observing $f_p$ with the best (unknown {\it a priori}) option is $\psi^*$, the cumulative regret is written as below \cite{auer2002}, 
\begin{align} \label{eq: cumulative_regrets}
    \mbox{Regret}(T) = T\psi^* - \sum_{k=1}^K N_T(k) \psi_k,
\end{align}
where $T$ is the total number of decision instances, $N_T(k)$ represents the number of times an option $k$ is selected within $T$ iterations, and $\psi_k$ is the probability that $f_p$ is observed by selecting the option $k$. In order to minimize the cumulative regrets, the robot is required to efficiently estimate the set of softmax parameters $\vec{\Theta}_k$ for all $k$ in the process of finding an optimal option by iteratively performing the two steps of option selection and measurement update. As described in Sec. \ref{sec: cmabs}, for measurement update, Bayes' theorem is commonly used. For option selection, $\varepsilon$-greedy, methods based on the upper confidence bound (UCB) and the softmax function are widely used \cite{kuleshov2014}. However, these methods cannot leverage additional information regarding the preference of observed outcomes, which could enable the robot to selectively favor options, leading to preferred outcomes and potential increase of the interpretability of the robot (issue A). Additionally, since these methods rely on heuristics for exploring the unknown options, they usually require lots of decision instances to determine the optimal option, which is not desirable for problems for which there are constraints on $T$ (issue B). Hence, in this study, we employ an option selection method that not just {\it exploits} the outcome preference to increase the interpretability of robotic decision-making, but also {\it explores} unknown options in a mathematically rigorous way for efficiently identifying the optimal options.   

\subsection{Active Inference Option Selection}
As experimentally validated in previous studies \cite{markovic2021, wakayamaICRA2023, wakayama2023observationaugmented}, option selection based on active inference (AIF) addresses the aforementioned desirable key elements. This is because of the unique characteristics of its objective function, 
i.e. expected free energy (EFE), which is composed of (i) the {\it extrinsic} value scoring the degree of how the predicted outcome observation distribution aligns with the desired distribution, and (ii) the {\it epistemic} value evaluating how executing an option could reduce the uncertainty of the option \cite{friston2015active}. In the following, we begin with outlining the derivation of EFE for constructing an option selection policy. Then, as a special case, we explain how to compute EFE when a prior proposal distribution for a hidden linear parameter vector $\vec{\Theta}$
is a multivariate Gaussian and the observation likelihood is the softmax function.

\subsubsection{Derivation of Option Selection Policy in AIF} According to the theory of active inference \cite{smith2022step, parr2022}, the goal of a decision-making agent is to minimize the {\it surprise} of 
observations to maintain its homeostasis. The surprise in the case of CMABs defined in Sec. \ref{sec: problem_statement} is expressed as, 
\begin{align} \label{eq: surprise}
    \mbox{Surprise} = -\log p(o) = -\log \int_{\vec{\Theta}} p(o, \vec{\Theta}) d\vec{\Theta}.
\end{align}
However, directly calculating (\ref{eq: surprise}) via multiple integrals tends to be analytically intractable, so its upper bound derived from Jensen's inequality results in a function called {\it free energy} (a.k.a. (negative) evidence lower bound) which is minimized instead. Nevertheless, in decision making, outcomes $o$ are unknown until an option $k$ is actually executed. Therefore, the AIF agent instead optimizes EFE (denoted as $G(a_k)$) described in \eqref{eq: efe}. Hereafter, the decision instance index $t$ and the context vector $\vec{x}_{k,t}$ are abbreviated for the ease of notation,  
{\allowdisplaybreaks
\abovedisplayskip = 3pt
\abovedisplayshortskip = 3pt
\belowdisplayskip = 3pt
\belowdisplayshortskip = 3pt
\begin{align} \label{eq: efe}
G(a_k)\!&=\!\int_{\vec{\Theta}_k}\!q(\vec{\Theta}_k|a_k)\!\sum_o p(o|\vec{\Theta}_k)\!\log\!\frac{q(\vec{\Theta}_k|a_k)}{p(\vec{\Theta}_k|o,\!a_k) p_{\mathrm{pr}}(o)} d\vec{\Theta}_k, \nonumber \\
&= \sum_o \Big\{ q(o|a_k) \log \frac{q(o|a_k)}{p_{\mathrm{pr}}(o)} \nonumber \\ 
&\qquad -\int_{\vec{\Theta}_k} q(\vec{\Theta}_k|a_k)p(o|\vec{\Theta}_k) \log p(o|\vec{\Theta}_k) d\vec{\Theta}_k \Big\},
\end{align}
}where $q(\vec{\Theta}_k|a_k)$ is a proposal distribution that approximates the posterior distribution $p(\vec{\Theta}_k|o, a_k)$, and $p_{\mathrm{pr}}(o)$ is a prior preference distribution, which defines an outcome observation distribution that the agent expects to see when undertaking options. Since $p_{\mathrm{pr}(o)}$ can be arbitrarly determined, in the case of the mineralogical survey scenario, for example, a human scientist can provide the robot with the desired mineral label distribution as $p_{\mathrm{pr}(o)}$, specified as $1\times F$ probability vector with non-negative entries summing to $1$. This desired distribution can be interpreted as a probabilistic characterization of worthwhile data that the scientist would expect to obtain. Specifying this distribution differentiates AIF from other conventional decision-making algorithms \cite{kaelbling1998, kurniawati2022}, where either robotics experts must manually adjust numeric reward values assigned to actions (a process that lacks straightforward interpretability), or rewards must be learned from multiple user demonstrations (which is also time-consuming and impractical for many kinds of exploration missions). In AIF literature, (\ref{eq: efe}) is commonly further transformed as follows to easily interpret the meaning,
\begin{align} \label{eq: efe_meaning}
   G(a_k) = &- \mathbb{E}_{q(o|a_k)}\Big[\log p_{\mathrm{pr}}(o)\Big]  \nonumber \\
    &-\mathbb{E}_{q(o|a_k)}\Big[D_{KL}\Big(q(\vec{\Theta}_k|o,a_k)||q(\vec{\Theta}_k|a_k)\Big)\Big], 
\end{align}
where $q(o|a_k)$ is the predicted observation distribution
\begin{align} \label{eq: predicted_observation}
    q(o|a_k) = \int_{\vec{\Theta}_k} q(\vec{\Theta}_k|a_k) p(o|\vec{\Theta}_k) d\vec{\Theta}_k.
\end{align}
The first term and the second term of \eqref{eq: efe_meaning} represent (i) the (negative) extrinsic value and (ii) the (negative) epistemic value, respectively. Thus, as can be seen from this equation, by optimizing (i.e. minimizing) $G(a_k)$, the agent can naturally strike a balance between {\it exploitation} contributing to (issue A) and {\it exploration} contributing to (issue B).
For detailed equation transformations to obtain \eqref{eq: efe} and \eqref{eq: efe_meaning}, refer to previous studies \cite{smith2022step, wakayamaICRA2023}. 

To further reflect the possibility that the agent is not necessarily confident of the values of $G(a_k)$, 
in this study, an option selection policy \eqref{eq: stochastic_policy} is formed with the use of $G(a_k)$, such that the agent samples the next action from the categorical distribution \eqref{eq: sample_index}. 
\begin{align} \label{eq: stochastic_policy}
        q(a_k) = \frac{\exp(-\gamma G(a_k))}{\sum_{j=1}^K \exp (- \gamma G(a_j))},
\end{align}
\begin{align} \label{eq: sample_index}
    a \sim \mbox{Cat}(a_1, \cdots, a_K; q(a_1), \cdots, q(a_K)). 
\end{align}
In \eqref{eq: stochastic_policy}, $\gamma$ is called {\it precision} (similar to inverse temperature) and it adjusts the confidence of the current EFE prediction \cite{smith2022step} (the larger the value of $\gamma$, the higher the confidence). Algorithm \ref{alg: aif_cmab} summarizes the process of active inference option selection for CMABs. At first glance, this stochastic option selection policy resembles the softmax option selection technique used for  conventional MABs and CMABs \cite{kuleshov2014}. However, unlike AIF, the conventional MAB softmax method {\it only} uses the average reward/utility obtained by selecting an option $k$ up until the current decision instance to calculate the probability of selecting that option. In other words, it does not take into account the prediction of future outcomes by utilizing context information as well as the (human) prior preference regarding outcome observations. The differences of the behaviors between softmax and AIF agents are further discussed in Sec. \ref{sec: simulation_study}. 

\begin{algorithm}[t] 
\caption{Active Inference Option Selection for CMABs} \label{alg: aif_cmab}
\begin{algorithmic}[1] 
\renewcommand{\algorithmicrequire}{\textbf{Input:}}
\renewcommand{\algorithmicensure}{\textbf{Output:}}
\REQUIRE Estimated set of parameters $\vec{\Theta}_k$ and context vector $\vec{x}_{k,t}$ for all options $k, k = \{1, \cdots, K\}$, and the prior preference distribution $p_{\mathrm{pr}}(o)$
\ENSURE Selected option index
\STATE Initialize $G(a_k)$ for all options
\FOR {each option $k$}
\FOR {each outcome $o$}
\STATE Compute $G(a_k, o)$ via (\ref{eq: efe_meaning})
\ENDFOR
\STATE Derive $G(a_k) = \sum_o G(a_k, o)$
\ENDFOR
\STATE Construct the option selection policy Cat$(\cdot)$ via (\ref{eq: stochastic_policy}) 
\RETURN Sample the option $a$ from (\ref{eq: sample_index})
\end{algorithmic} 
\end{algorithm} 

\subsubsection{Special Case: Multivariate Gaussian Prior and Softmax Observation Likelihood} 
When  
$q(\vec{\Theta}_k|a_k)$ is multivariate Gaussian and 
$p(o|\vec{\Theta}_k)$ is a softmax function, $G(a_k)$ cannot be computed analytically since calculating (\ref{eq: predicted_observation}) is intractable. Luckily, several statistical methods have been proposed to approximate this normalization term \cite{nisarTRO, wakayamaTRO, bishop2006}, and in this study we adopt the Laplace approximation due to its computation efficiency.

In statistical machine learning, the Laplace approximation is often employed to approximate a probability density function (pdf) as a Gaussian distribution \cite{bishop2006}. This uses the second-order approximation of the vector Taylor expansion of a logarithmic function whose gradient is a zero-vector. Particularly, in the process of approximating $G(a_k)$, a function $g(\vec{\Theta}_k)$ is defined as the joint unnormalized distribution $q(\vec{\Theta}_k|a_k)p(o|\vec{\Theta}_k)$ such that the following logarithmic function is used, 
\begin{align}
    \log g(\vec{\Theta}_k) &\approx \log g(\vec{\Theta}_{k}^{(0)}) \nonumber \\ 
    &+ \sum_{r=1}^{(C+1)\times F} (\Theta_{k,r} - \Theta_{k,r}^{(0)})\frac{\partial \log g(\vec{\Theta}_k^{(0)})}{\partial \Theta_{k,r}} \nonumber \\
    &+ \frac{1}{2} \Big\{\sum_{r=1}^{(C+1)\times F} (\Theta_{k,r} - \Theta_{k,r}^{(0)}) \frac{\partial \log g(\vec{\Theta}_{k}^{(0)})}{\partial \Theta_{k,r}} \Big\}^2, \label{eq: second_order_approximation}
\end{align}
where $\vec{\Theta}_k^{(0)}$ satisfies $\nabla \log g(\vec{\Theta}_k) = \vec{0}$. However, as it is also analytically intractable to find $\vec{\Theta}_k^{(0)}$, $\vec{\Theta}_{k, MAP}$ is computed via Newton's method \cite{galantai2000}. Since the second term in \eqref{eq: second_order_approximation} is removed and the third term in (\ref{eq: second_order_approximation}) can be written as
\begin{align}
    \frac{1}{2} \Big\{\sum_{r=1}^{(C+1)\times F} & (\Theta_{k,r} - \Theta_{k,r}^{(0)}) \frac{\partial \log g(\vec{\Theta}_{k}^{(0)})}{\partial \Theta_{k,r}} \Big\}^2 \nonumber \\
    &= \frac{1}{2} \Big\{ (\vec{\Theta}_k - \vec{\Theta}_k^{(0)})^T \nabla \log g(\vec{\Theta}_k^{(0)}) \Big\}^2, 
\end{align}
(\ref{eq: second_order_approximation}) reduces to 
\begin{align}
    &\log g(\vec{\Theta}_k) \approx \log g(\vec{\Theta}_{k, MAP}) + \nonumber \\ 
    &\frac{1}{2} (\vec{\Theta}_k\!-\!\vec{\Theta}_{k, MAP})^T \mbox{H}\Big[\!\log g(\vec{\Theta}_{k, MAP})\!\Big] (\vec{\Theta}_k\!-\!\vec{\Theta}_{k, MAP}), \label{eq: laplace_with_hessian}
\end{align}
where H is the Hessian (Note that the Hessian can be calculated by computing the Jacobian of the gradient, and the gradient can be derived with an optimizer implemented in the standard scientific computing library such as {\tt scipy.optimizer}). Thus, if we define $A = - $H and by taking the logarithm from (\ref{eq: laplace_with_hessian}), 
\begin{align}
    &g(\vec{\Theta}_k) \approx \nonumber \\
    &g(\vec{\Theta}_{k,MAP})\!\cdot\!\mbox{exp} \Big(\!-\!\frac{(\vec{\Theta}_k\!-\!\vec{\Theta}_{k,MAP})^T A (\vec{\Theta}_k\!-\!\vec{\Theta}_{k,MAP})}{2}\Big), 
\end{align}and the normalization constant, i.e. the predicted observation distribution \eqref{eq: predicted_observation}, is computed as
\begin{equation} \label{eq: predicted_observation_analytical}
    q(o|a_k) = g(\vec{\Theta}_{k,MAP}) \cdot\frac{(2\pi)^{\frac{(C+1)\times F}{2}}}{|A|^{\frac{1}{2}}}.
\end{equation}By using \eqref{eq: predicted_observation_analytical} into \eqref{eq: efe}, the first term of \eqref{eq: efe} (i.e. $q(o|a_k) \log \frac{q(o|a_k)}{p_{\mathrm{pr}}(o)}$) can be calculated. To calculate the second term of \eqref{eq: efe}, by approximating $p(o|\vec{\Theta}_k)$ as a Gaussian exponential form from the result of the Laplace posterior approximation. More details can be found in \cite{wakayama2023observationaugmented}.

\section{Simulation Study} \label{sec: simulation_study}
To verify whether the proposed active inference option selection method is effective not only for stationary, independent, and linear CMABs formulated with randomly generated hidden parameters and contexts, as in previous studies \cite{wakayamaICRA2023, wakayama2023observationaugmented}, but also for CMABs formulated based on actual scientific data, a mineral search site selection study is considered. In the following subsections, we begin with an overview of the motivating autonomous robotic exploration scenario focusing on surface mineralogical surveys. We then describe the hyperspectral and mineral label dataset used as contexts and outcome observations. This is followed by an explanation of the preprocessing steps and the result of learning the true hidden parameters necessary for calculating the cumulative regret. Finally, we detail the simulation setup and present the results of Monte Carlo simulation experiments under both static and dynamic human prior preferences. 

\subsection{Motivating Scenario} \label{sec: motivating_scenario}
Limestone and iron are indispensable in construction and manufacturing sectors. Minerals such as kaolinite and pyroxene play a crucial scientific role, shedding light on sedimentary processes and enhancing our comprehension of rock formation \cite{sabins1999remote}. Consequently, mineralogical surveys in unfamiliar territories are pivotal for uncovering resources and driving scientific advancements. Nevertheless, the extensive scope of mineral exploration presents time and safety challenges, making effective human-led surveys difficult. As a result, research has been conducted to utilize robots equipped with sensing suits to autonomously perform exploration \cite{candela2017, arora2019}. 

In the following, we consider the problem of an autonomous aerial robot (as shown in Fig. \ref{fig: scenario}) identifying the most promising site(s) where a mineral rock specimen desired by a scientist can be sampled in a follow-up sample-return mission \cite{muirhead2020mars}. These $K$ number of sites are predetermined based on satellite images \cite{zurek2007overview}. In this problem, the aerial robot uses relatively lightweight sensors, such as a spectrometer, to scan the search sites, and predicts the site with the highest likelihood of containing the desirable specimen based on the obtained contextual hyperspectral information $\vec{x}$. The robot then receives an observation $f$ on the detected mineral\footnote{In this study, it is assumed that the total number of minerals present in the entire region is bounded by a finite value $F$ as with \cite{candela2021using}.} at the selected site $k$ from another robot, which is remotely operated by humans and can quickly access the scanned coordinate. By hierarchically structuring the search process in multiple stages as such, rather than exhaustively dispatching the robots to survey the entire region, it is expected that survey efficiency significantly improves. However, the outcome observation is probabilistic by nature and the latent relationship between the context $\vec{x}$ and the observation $f$ used to predict the likelihood of observing each mineral specimen are unknown {\it a priori}, so a CMAB described in Sec. \ref{sec: methodology} is adopted to carefully take a balance between exploitation and exploration.  

\subsection{Dataset} \label{sec: dataset}
\begin{figure}[t]
    \centering
    \includegraphics[width=0.9\linewidth]{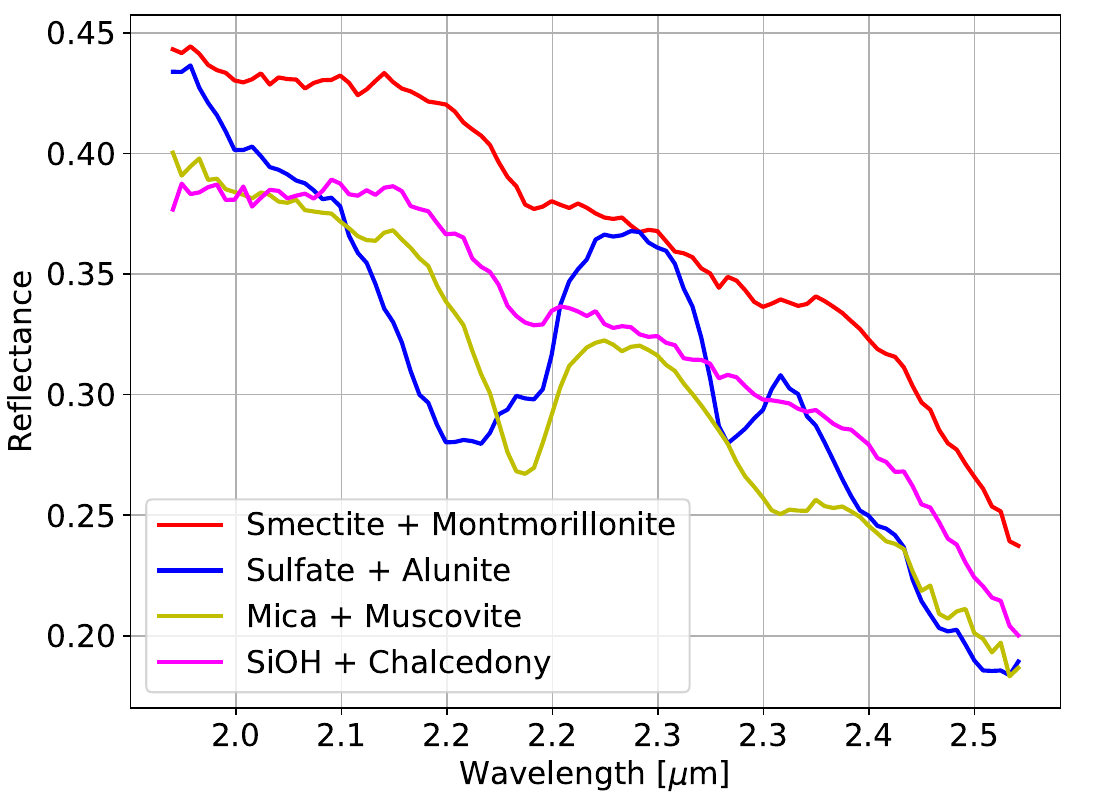}
    \captionsetup{skip=5pt}
    \caption{Example raw hyperspectral data from the AVIRIS-NG dataset.}
    \label{fig: raw_hyperspectral}
\end{figure}

The hyperspectral data used in this study is collected using the Next Generation Airborne Visible-Infrared Imaging Spectrometer (AVIRIS-NG) at the Cuprite mining district, Nevada, an area known for its high mineralogical diversity \cite{hamlin2011}. This data assigns a unique reflectance spectrum across $97$ spectral bands to every location (pixel) in the scene. Fig. \ref{fig: raw_hyperspectral} shows the example spectra collected at several different locations. On the other hand, the mineral label data is constructed by geological experts and one of $215$ labels are assigned to each pixel \cite{swayze2014mapping}. Fig. \ref{fig: mineral_map} represents the mineral map highlighted with arbitrary colors. Note that these two maps are aligned so that the sizes of map images ($2673$ and $2389$ pixels in height and width directions) and pixels ($3.9$ m square) are the same.

\subsection{Training True Latent Parameters} \label{sec: train_params}
When calculating the cumulative regret to evaluate and comparing the performance of option selection algorithms, the ground truth best-fit softmax parameters $\vec{\Theta}^{*}_k$ are required to sample the outcomes for the best possible case. Note that these softmax parameters are never known by a decision-making agent during deployment and can {\it only} be accessed/trained offline (i.e. one of the goals of the decision-making agent is to efficiently learn the values of these parameters). In this subsection, we outline the preprocessing steps for the hyperspectral and mineral dataset introduced in Sec. \ref{sec: dataset} and detail the training procedure of the ground truth best-fit softmax parameters\footnote{Note that the true underlying statistics are not necessarily represented by a linear softmax model. This modely simply represents the best approximation the autonomous robot could achieve using a linear approach, assuming it had access to more data and ground truth labels.}. 

\begin{figure}[t]
    \centering
    \includegraphics[width=0.65\linewidth]{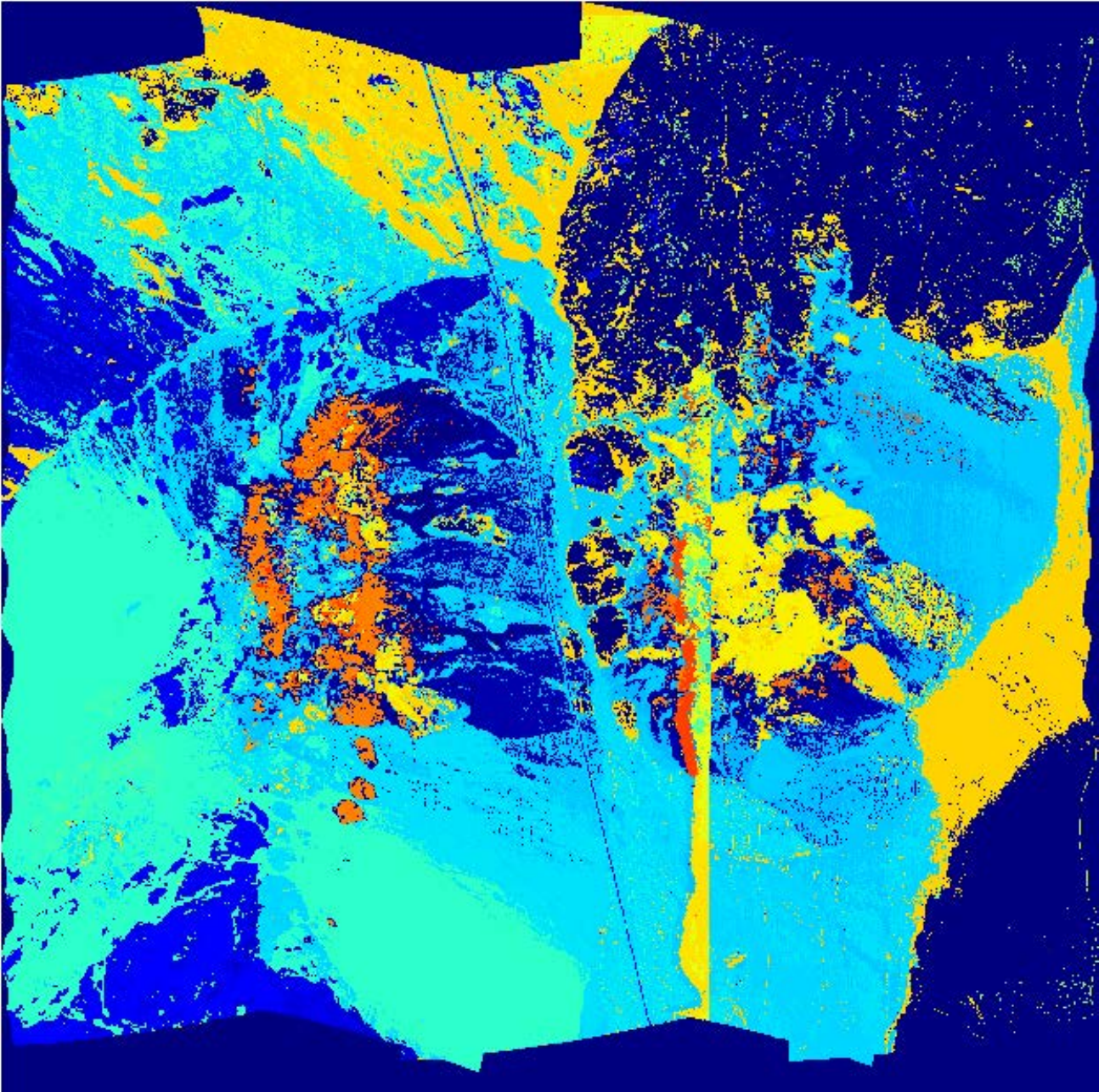}
    \captionsetup{skip=5pt}
    \caption{Mineral map: different colors correspond to different (mixture) mineral labels. In total, there are $215$ labels in this region. Note that pixels on the edge with dark blue colors are invalid and no mineral labels are assigned. These pixels are ignored when training true softmax parameters.}
    \label{fig: mineral_map}
\end{figure}

\begin{figure}[t]
    \centering
    \includegraphics[width=0.9\linewidth]{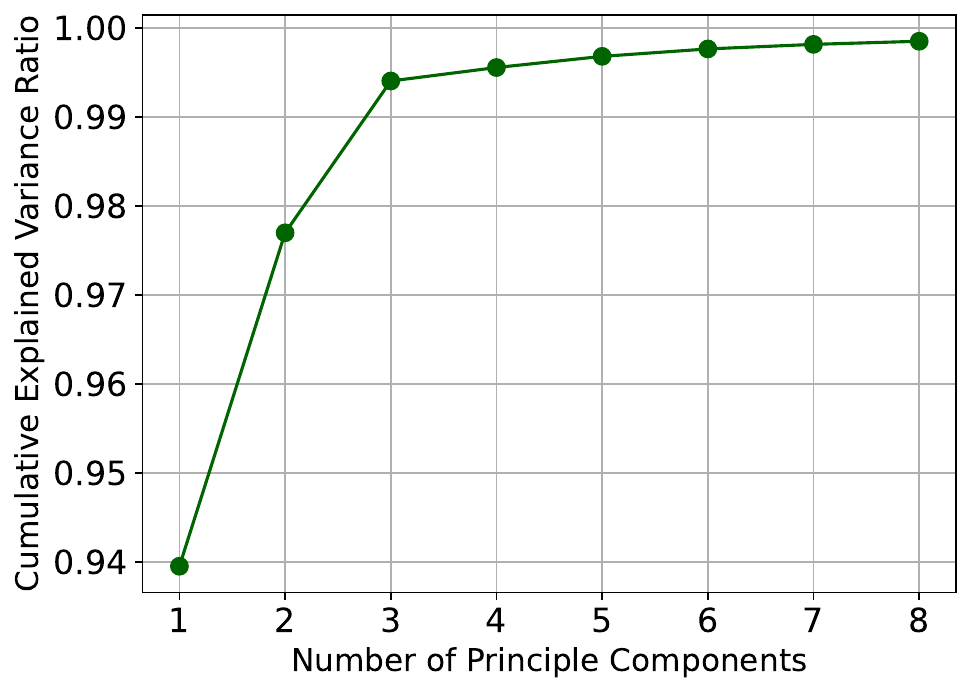}
    \captionsetup{skip=5pt}
    \caption{Transition of the cumulative explained variance ratio after applying PCA to the AVIRIS-NG dataset.}
    \label{fig: pca_result}
\end{figure}

First of all, some pixels lack hyperspectral data, while others lack mineral label data. Since these pixels do not necessarily overlap, we take the union of these sets, marked them as invalid pixels (shown in dark blue in Fig. \ref{fig: mineral_map}), and excluded them from the training process. Next, since the AVIRIS-NG dataset has a very high spectral resolution, its dimensionality $C$ is reduced from $97$ to $8$ via principal component analysis (PCA) \cite{bishop2006}. This dimensionality reduction is plausible as the cumulative explained variance ratio (i.e. the sum of the target number of eigenvalues divided by the sum of all eigenvalues, which ranges between $0$ and $1$) when the number of PCA components is $8$ is $0.999$ (Fig. \ref{fig: pca_result}). Additionally, since several mixtures of minerals assigned with different labels are quite similar and some labels are not actually used, the mineral label dataset is further manually clustered from $215$ to $14$ with the advice of experts. Exemplary representative minerals in these 14 clusters include alunite, mica, and kaolinite. 
\begin{figure}[t]
    \centering
    \includegraphics[width=0.8\linewidth]{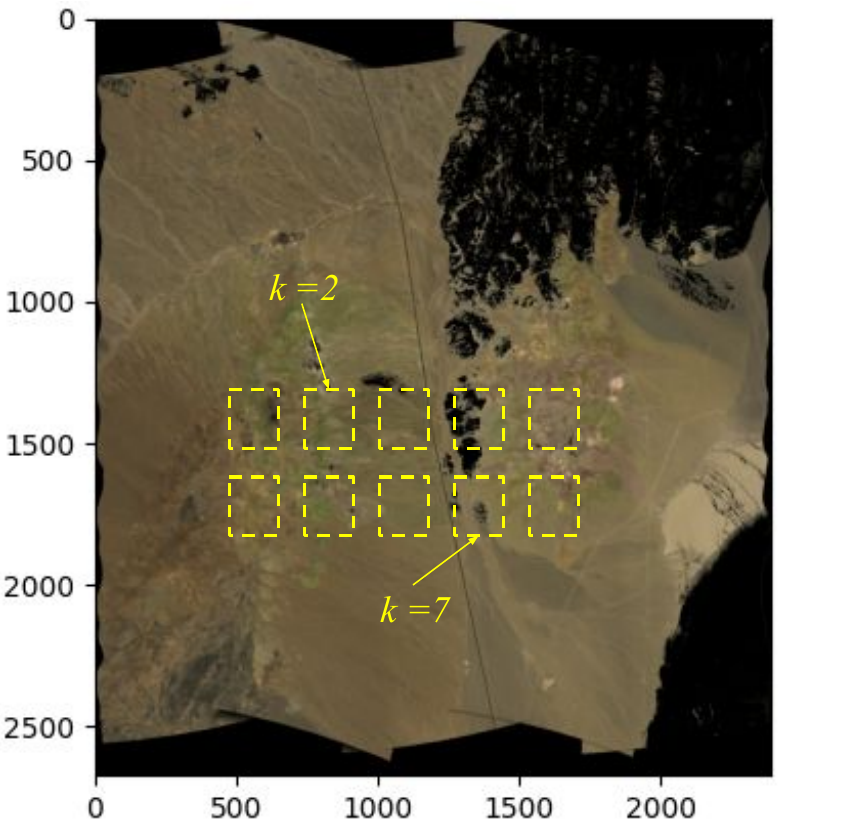}
    \captionsetup{skip=5pt}
    \caption{Selected search sites overlaid on an aerial image of the Cuprite mining district, Nevada.
    }
    \label{fig: entire_region}
\end{figure}After these preprocessing steps, as shown in Fig. \ref{fig: entire_region}, $K$ non-overlapping search sites are selected, each with dimensions of $200$ pixels in width and $250$ pixels in height, and the pairs of hyperspectral and mineral label data are combined as datasets. To train the true best-fit softmax parameters $\vec{\Theta}^{*}_k$, the dataset is split into training (80\%) and test sets (20\%) and the softmax regression (i.e. multinomial logistic regression) with the Adam optimizer \cite{kingma2014adam} is performed with PyTorch \cite{paszke2019pytorch}. The average accuracy (i.e. the proportion of correctly classified samples out of the total number of samples) over all search sites is $76.7\%$ (note that min/max accuracy is $62.3\%$ and $93.1\%$, respectively). Despite having a relatively low accuracy as a classifier, it effectively captures the noise present in the measurement process as shown in Fig. \ref{fig: confusion_matrix}. This is further confirmed by examining the histograms of the learned bias values. As depicted in Fig. \ref{fig: histograms_biases}, each subplot exhibits significant negative values (around $-20$). This observation indicates that the trained classifier discerns the absence of certain minerals in these search sites\footnote{For example, in Fig. \ref{fig: confusion_matrix}, it can be observed that minerals of Classes $2$ and $13$ are absent at sites $3$ and $6$. As the number of such absent minerals increases, the frequency of large negative values in Fig. \ref{fig: histograms_biases} also increases.}. Additionally, considering that other research utilizing a similar dataset also demonstrates comparable accuracy values, it suggests that this classifier is satisfactory to provide a best fit baseline comparison \cite{candela2017}.
\begin{figure}[t]
    \centering
    \includegraphics[width=\linewidth]{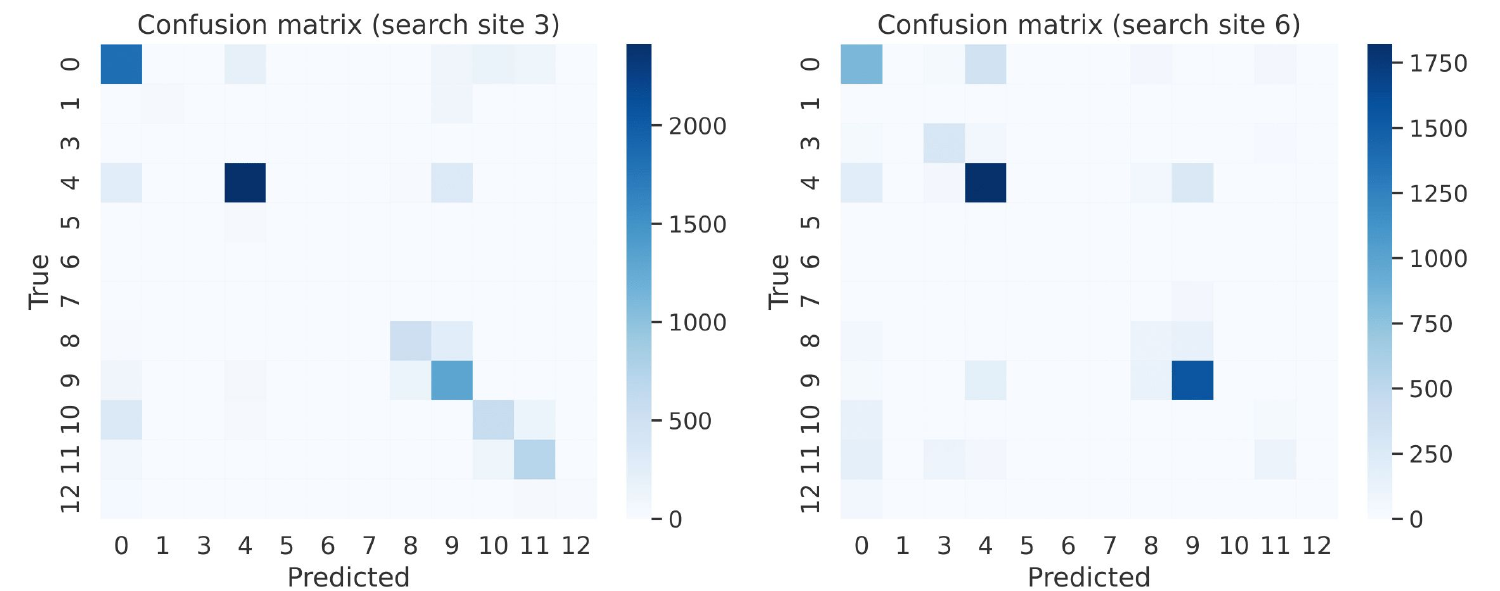}
    \caption{Examples of the confusion matrices constructed from the test results.}
    \label{fig: confusion_matrix}
\end{figure}

\begin{figure}[t]
    \centering
    \includegraphics[width=\linewidth]{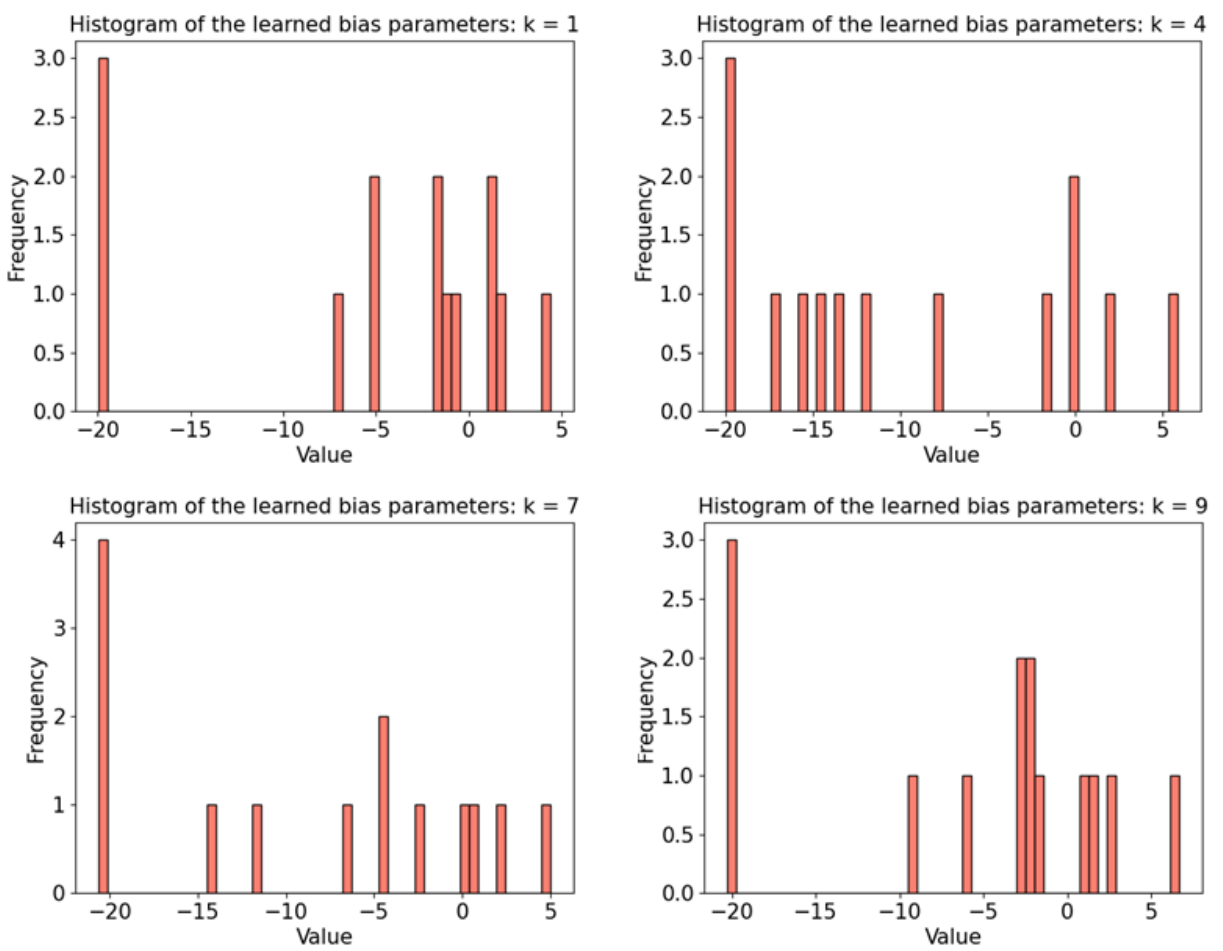}
    \caption{Examples histograms of the learned bias values; in mathematical terms, when a bias term $b_{k,f}$ is highly negative, the numerator in the softmax likelihood function associated with this label $f$, i.e. $e^{\vec{w}^T_{k,f} \vec{x}_{k,t} + b_{k,f}}$, tends towards zero, causing $p(o_{k,t} = f|\vec{\Theta}_k; \vec{x}_{k,t})$ to also approach zero.}
    \label{fig: histograms_biases}
\end{figure}

\subsection{Simulation Setup}
\begin{figure}[t]
    \centering
    \includegraphics[width=\linewidth]{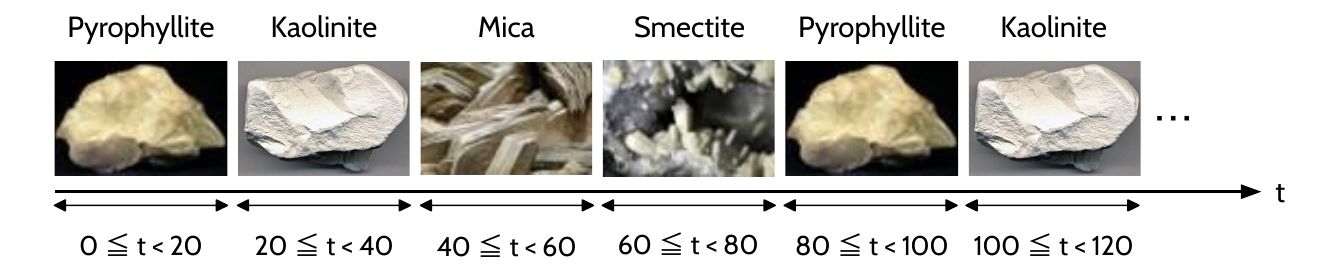}
    \caption{Transition of the mineral of greatest interest to a scientist. In this study, for simplicity, it is assumed that transitions occur every $20$ instances.}
    \label{fig: dynamic_prior_preference}
\end{figure}

With the set of the trained ground truth softmax parameters, the following option selection methods are considered and compared in extensive Monte Carlo (MC) simulation: (i) best-fit optimal option selection, using the trained parameters (required for computing the cumulative regret); (ii) $\varepsilon$-greedy (where $\varepsilon\!=\!0.3$ was found to work best after initial trials); (iii) softmax method (where temperature $\tau$ was set as $0.1$ after initial trials); (iv) upper confidence bound (UCB) (where the exploration parameter $c$ was set as $0.8$ after initial trials); (v): multicategorical Thompson sampling (TS); (vi): active inference (AIF; where precision $\gamma$ was set as 30 after initial trials). The option selection methods (v) and (vi) are paired with the Laplace approximation for the measurement update \cite{bishop2006}. $100$ MC runs are performed, and the number of iterations $T$ in each MC run is set to $100/150$, which is much smaller compared to common MAB algorithm benchmarks \cite{markovic2021} and reflects a practical upper limit for robotic lander sensor deployment \cite{wakayama2023observationaugmented}. When the robot is actually deployed, the hyperspectral contextual information at each site varies across decision instances because the exact coordinates targeted by the spectrometer differ each time. To replicate this real-world stochasticity, at every decision-making instance, a pixel is randomly selected (corresponding to its coordinates) within each site, and the PCA-processed hyperspectrum associated with it is used as the context $\vec{x}_{k,t} \in \mathbb{R}^C$. For the initial probability distribution $p(\vec{\Theta})$ used to estimate the hidden softmax parameter vector, a multivariate normal distribution is employed across all search sites, with a mean vector where all elements are $0.5$ and a diagonal covariance matrix with a scaling factor of $5$. Note that the value of the scaling factor is determined after initial trials. Finally, in the first simulation experiment intended to validate the effectiveness of the proposed AIF-based option selection method in real scientific missions, it is assumed that a scientist holds the strongest and consistent/stationary interest in observing pyropillite specimens (i.e. $o=f_p$). Thus, the prior preference for observing pyropillite specimens $p_{ev}(o=f_p)$ is set to $0.8$, while $p_{ev}(o \neq f_p)$ is set to $0.2$ divided by the $13$ other possible outcomes. In contrast, the second experiment assumes that the minerals of greatest interest to scientists (often informed by insights gained up to that point) dynamically changes as shown in Fig. \ref{fig: dynamic_prior_preference} to better align with real scientific missions, and verifies how the proposed method adapts to this variability. 

\subsection{Results: Stationary Prior Preference}
\begin{figure}[t]
    \centering
    \includegraphics[width=\linewidth]{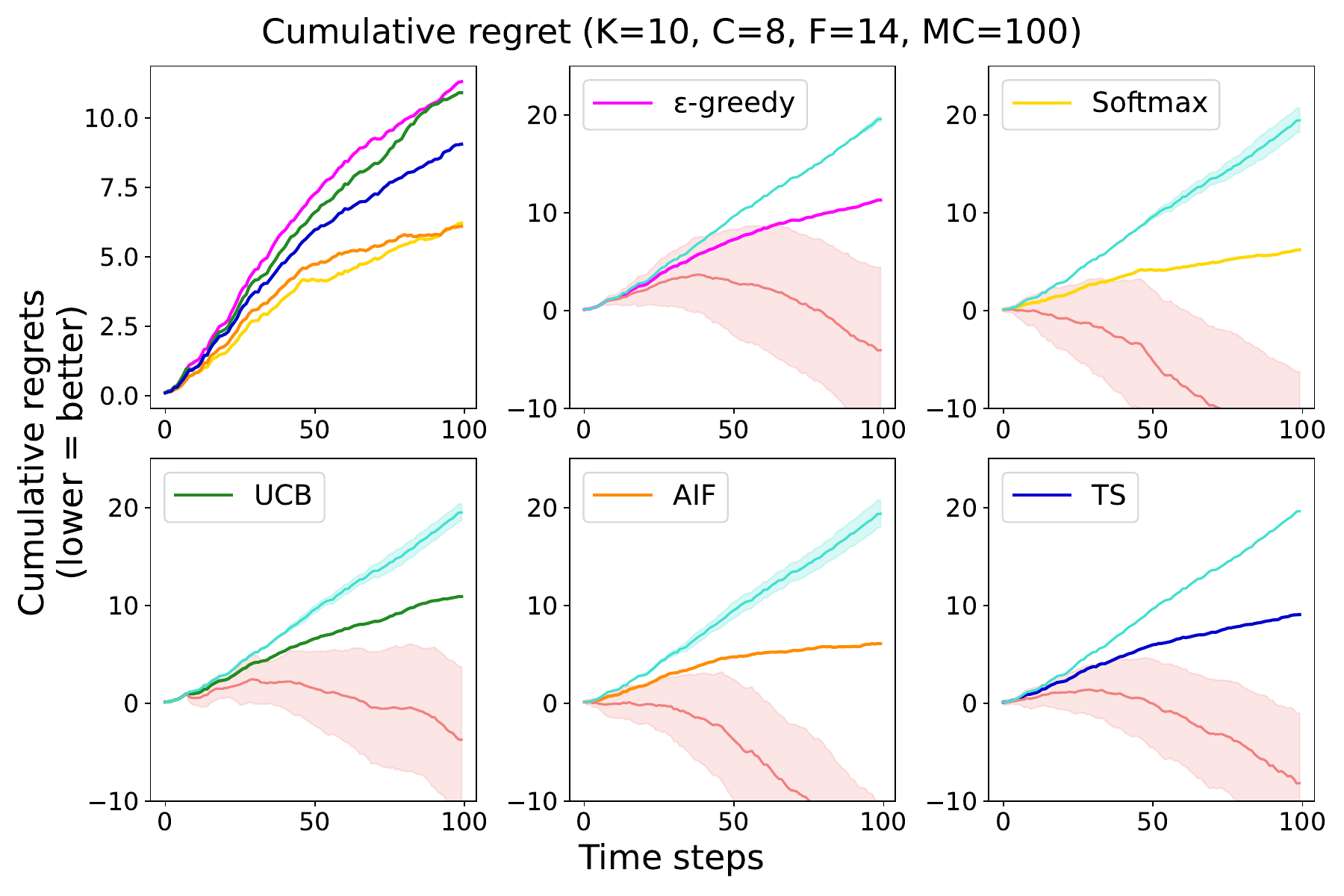}
    \caption{Comparison of the cumulative regrets when the prior preference is stationary (top left) and the cumulative regrets for each option selection method (others). In the subplots other than the top-left one, there are turquoise and salmon-colored lines and shaded regions. These represent the means and $1$-$\sigma$ bounds of the sets where the cumulative regret value at the final step is above and below the overall average, respectively.}
    \label{fig: cumulative_regret_stationary}
\end{figure}

\begin{figure}[t]
    \centering
    \includegraphics[width=\linewidth]{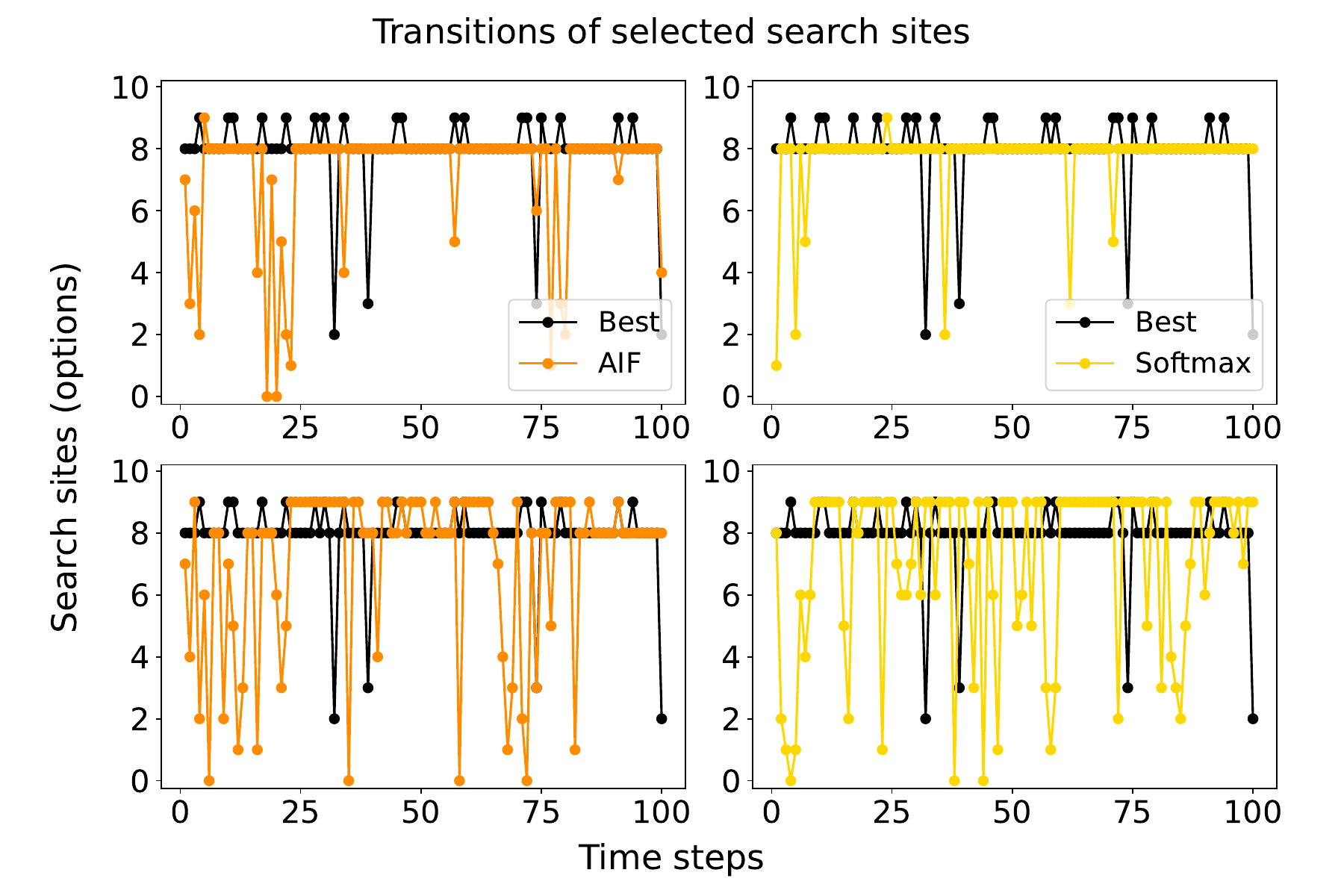}
    \caption{Example transitions of selected search sites when the AIF (left column) and softmax (right column) methods are used. The transitions shown in the top row result in very small cumulative regrets, while those in the bottom row lead to very high cumulative regrets.}
    \label{fig: option_transitions_good_bad}
\end{figure}

When the prior preference distribution is stationary, the cumulative regrets of both the proposed AIF method (orange) and the softmax method (yellow) outperform others as shown in Fig. \ref{fig: cumulative_regret_stationary} (upper left). 
Interestingly, in this case, there is notable variability in cumulative regrets across all methods as depicted in Fig. \ref{fig: cumulative_regret_stationary} (other subplots). Cumulative regret represents the difference between the ideal cumulative reward, assuming known hidden parameters, and the actual cumulative reward obtained from following a specific policy. As such, it generally remains non-negative. However, in this simulation study, during the preprocessing of the dataset, nonlinear transformations were applied, such as significantly reducing the total number of mineral labels used. Additionally, not all minerals were necessarily present at each site, and the test accuracy was not exceptionally high. Therefore, even when selecting sites based on the best-fit optimal option selection strategy, there is no guarantee that the obtained outcomes align with the outcome observation a scientist is interested in. Consequently, in some MC runs, alternative methods were found to yield lower cumulative regrets\footnote{In this simulation experiment, agents are stuck in local minima or continued exploring search sites throughout instances in approximately $35\%$ of the MC runs (corresponding to the turquoise lines).}. This type of bimodality in cumulative regrets can also be confirmed by observing the transitions of search sites selected by each method. As shown in the top row of Fig. \ref{fig: option_transitions_good_bad}, in one MC run, it can be seen that the AIF and softmax methods select the best search site (in this case, $k$=$8$) more frequently than when using the best possible option selection strategy. On the other hand, when stuck in local minima or continuing to explore the best search site, as shown in the bottom row, the frequency with which these methods select the best search site significantly decreases, resulting in higher cumulative regrets. 

\subsection{Results: Dynamic Prior Preference}
\begin{figure}[t]
    \centering
    \includegraphics[width=\linewidth]{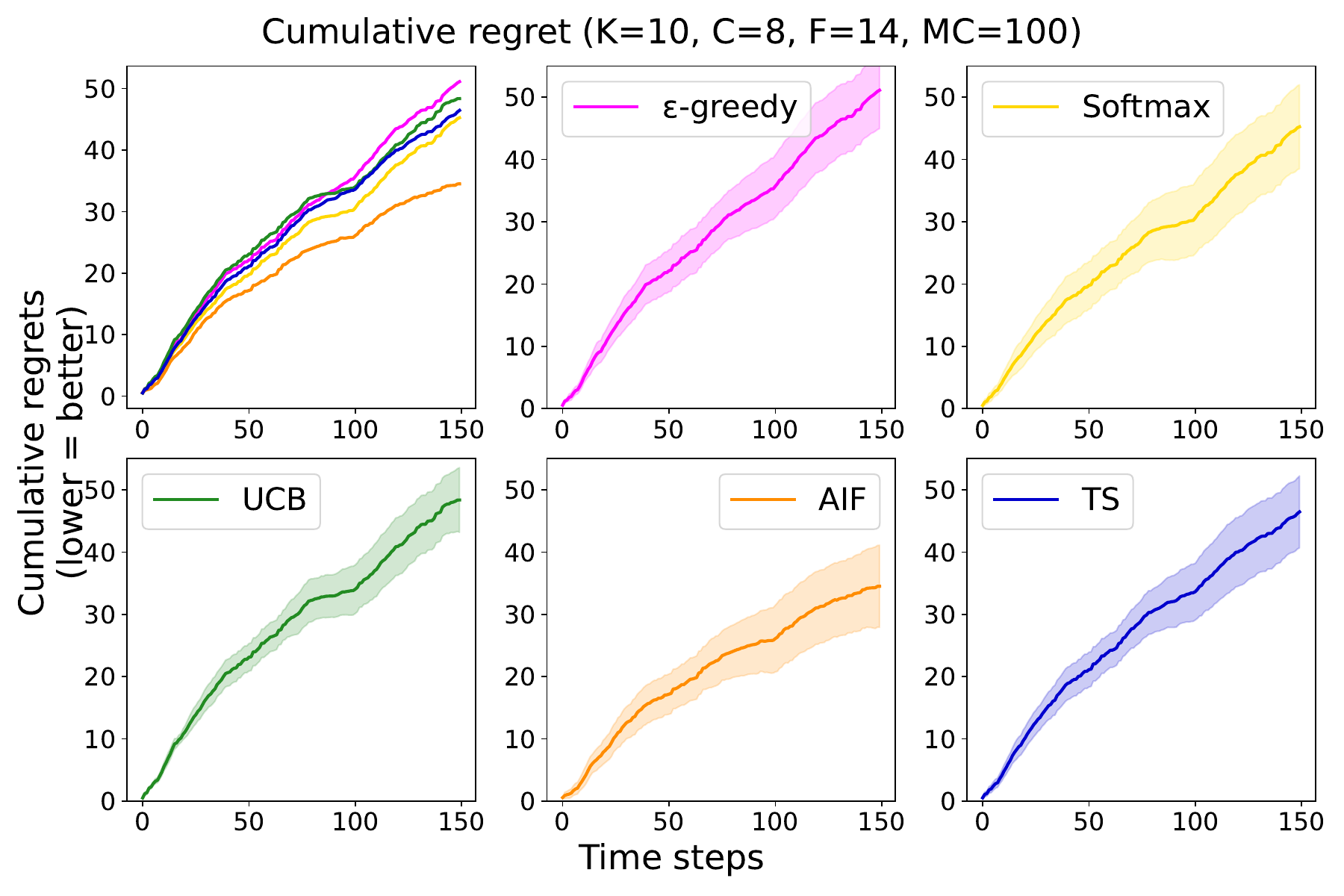}
    \caption{Comparison of the cumulative regrets when the prior preference is dynamically and periodically changed; the shaded regions represent the $1$-$\sigma$ bounds of cumulative regrets.}
    \label{fig: cumulative_regret_dynamic}
\end{figure}

\begin{figure}[t]
    \centering
    \includegraphics[width=\linewidth]{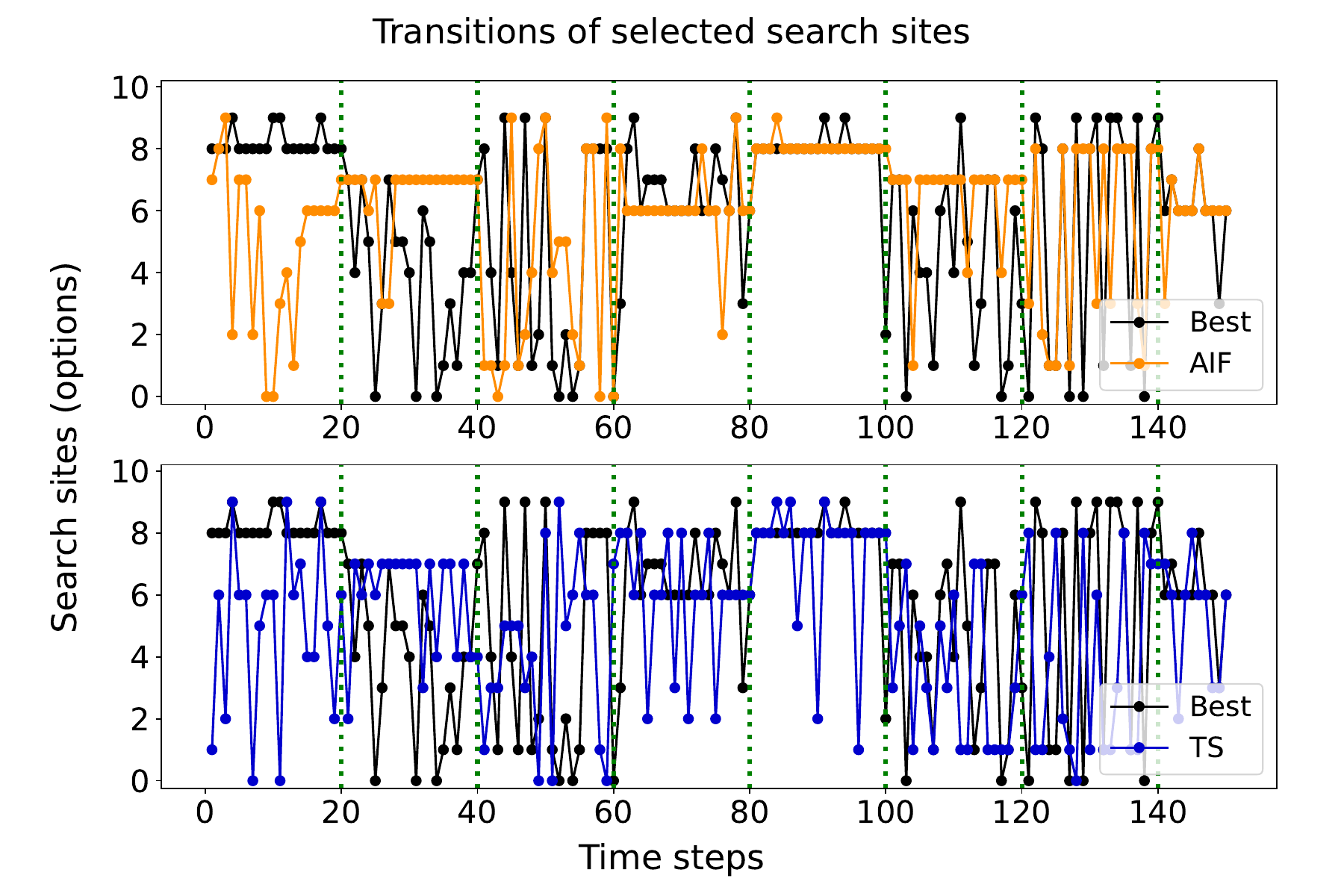}
    \caption{Example transitions of selected search sites when the AIF (top) and TS (bottom) are used. In this scenario, human prior preference changes every $20$ decision instances (green dotted lines).}
    \label{fig: option_transitions_dynamic}
\end{figure}

Fig. \ref{fig: cumulative_regret_dynamic} (top left) illustrates the comparison of the cumulative regrets when the prior preference distribution changes dynamically as shown in Fig. \ref{fig: dynamic_prior_preference}. In this scenario, the proposed AIF method demonstrates superior performance compared to all conventional option selection methods such as softmax and Thompson sampling. This superiority is stemmed from EFE's epistemic term efficiently assessing the uncertainties of search sites, thereby identifying sites with a high likelihood of achieving desired outcomes at each time step, even as scientists' desired observational outcomes change dynamically. For instance, as shown in Fig. \ref{fig: option_transitions_dynamic}, during the initial $20$ instances, neither AIF nor TS agents identify the site to observe pyrophillite. However, by the time when pyropyhillite becomes again a desired outcome (i.e. from $80$ to $100$ instances), the AIF agents are able to exploit the best site where pyrophylitte is likely to be observed, influenced by the extrinsic term. In contrast, the TS agents still continue to explore sites other than the best site. Additionally, in this simulation experiment, unlike when the prior preference is stationary, the significant variability in cumulative regrets is not observed. This is because the desired outcomes change regularly, so even if the accuracy of the trained hidden softmax parameters utilized in the best option selection strategy is not very high, using this allows for observing more desired outcomes compared to other strategies. 
%

\section{Conclusions} \label{sec: conclusion}
In this study, we applied active inference (AIF) as an option selection method for contextual multi-armed bandits (CMABs) with the objective of validating its efficacy using real scientific data. Previous studies primarily relied on synthetic data to simulate true hidden parameters and contexts. In contrast, we utilized actual hyperspectral data along with mineral labels for these values. Additionally, we detailed the preprocessing procedures and the methodology used to train the true hidden parameters of search sites. Our research comprised two sets of Monte Carlo simulation experiments. The first set primarily aimed to validate the effectiveness of the proposed AIF method under the assumption of stationary human prior preferences, consistent with prior studies. As a result, AIF agents demonstrated on par or superior performance compared to other existing option selection methods. In the second set of experiments, we introduced more realistic scenarios by assuming dynamic changes in human prior preferences. Interestingly, the proposed AIF method exhibited even greater performance improvements in these dynamic settings. This enhancement is attributed to the unique characteristics of expected free energy (EFE), which underpin AIF's ability to adapt and optimize exploration-exploitation tradeoffs efficiently in response to changing preferences.

\section*{Acknowledgments}
Work supported by the NASA COLDTech Program, grant \#80NSSC21K1031. S. Wakayama was also supported by the Masason Foundation. Part of this research was carried out at the Jet Propulsion Laboratory, California Institute of Technology, under a contract with the National Aeronautics and Space Administration (80NM0018D0004).

\bibliographystyle{IEEEtran}
\bibliography{root}


\begin{IEEEbiography}
[{\includegraphics[width=1in,height=1.25in,clip,keepaspectratio]{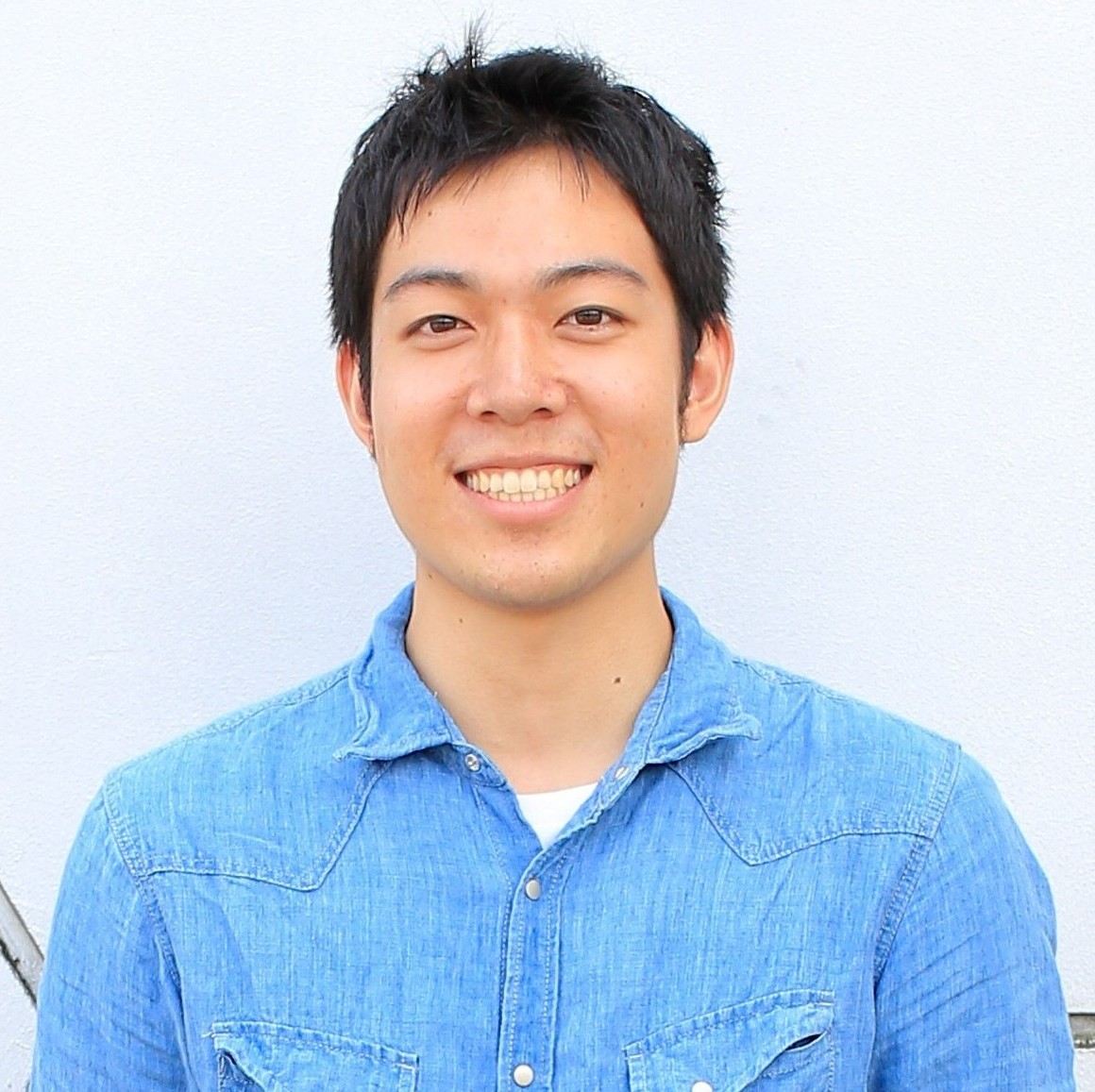}}]{Shohei Wakayama}
received the B.Eng. in Mechanical Engineering from Kyushu University in 2018, and the Ph.D. in Aerospace Engineering with the Ann and H.J. Smead Aerospace Engineering Sciences Department, University of Colorado Boulder in 2024. His research interests lie in Bayesian state estimation and sequential decision making under uncertainty, and human-robot interaction for robotic exploration of unknown remote environments.  
\end{IEEEbiography}

\begin{IEEEbiography}
[{\includegraphics[width=1in,height=1.25in,clip,keepaspectratio]{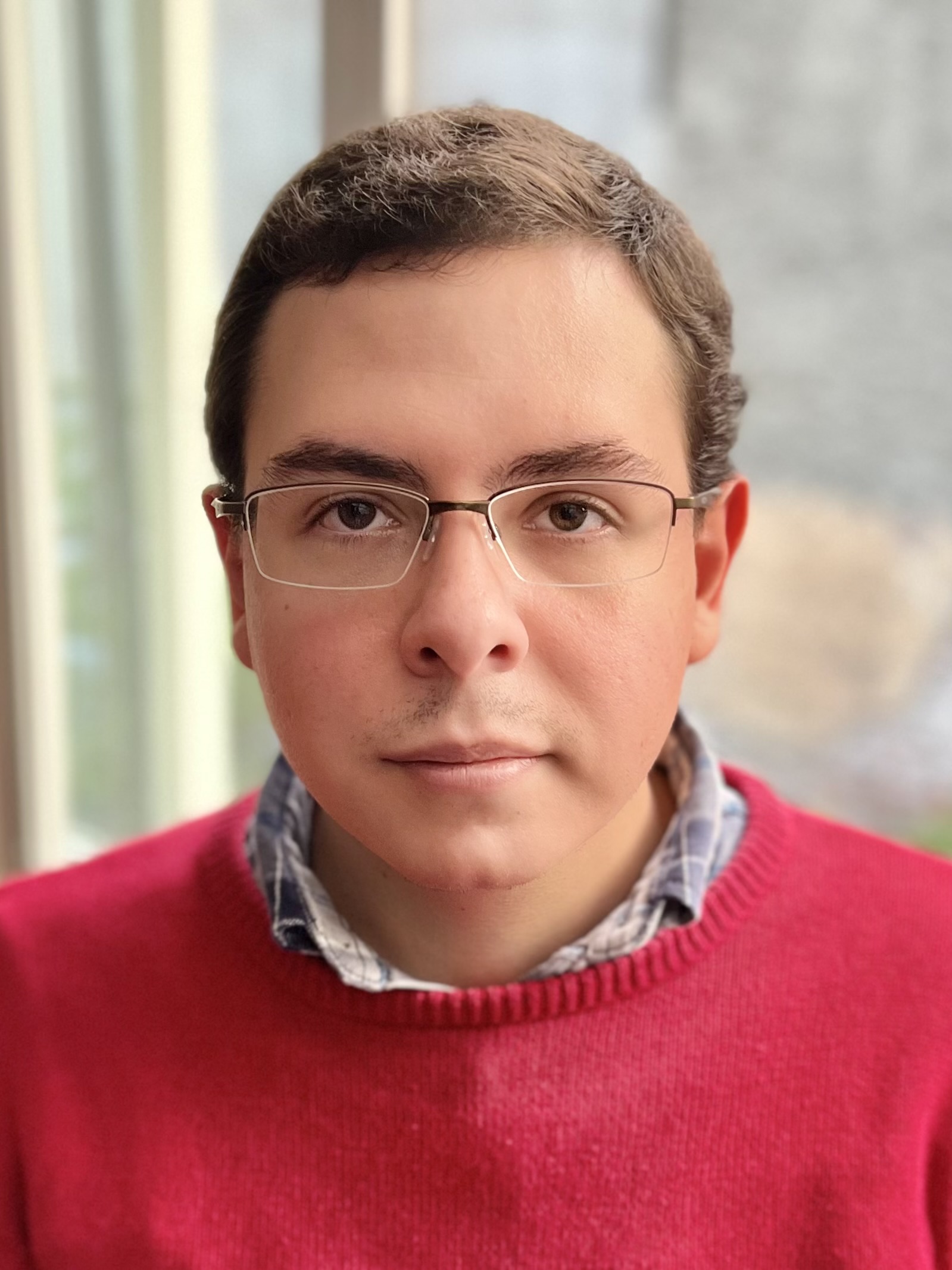}}]{Alberto Candela}
is a Data Scientist in the Artificial Intelligence Group at the Jet Propulsion Laboratory, California Institute of Technology. He received his B.S. in Mechatronics Engineering from Instituto Tecnológico Autónomo de México, and his M.S. and Ph.D. in Robotics from Carnegie Mellon University. His research interests include autonomous science, information-theoretic planning, machine and deep learning, probabilistic and statistical methods, robotics, and remote sensing.
\end{IEEEbiography}

\begin{IEEEbiography}
[{\includegraphics[width=1in,height=1.25in,clip,keepaspectratio]{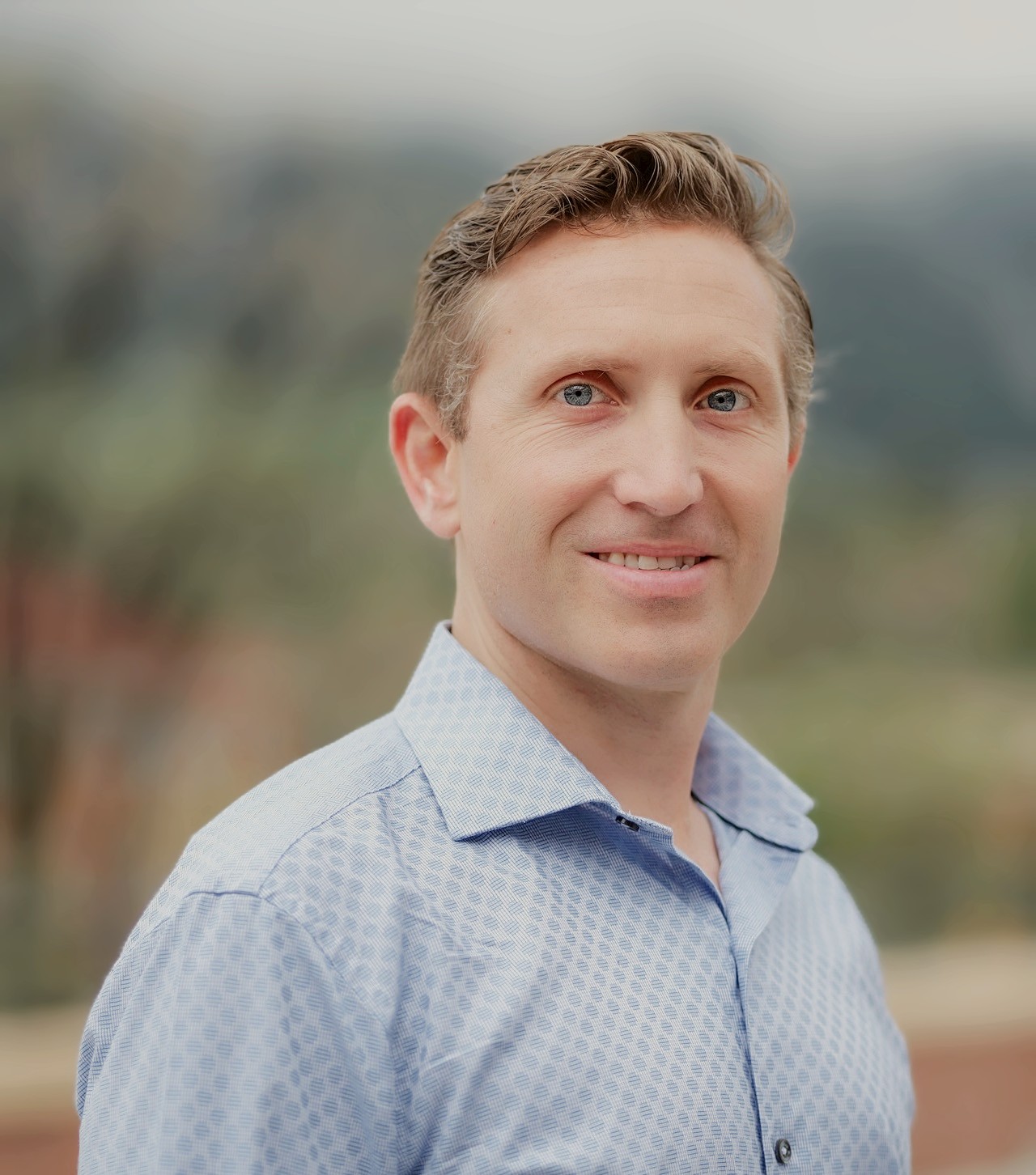}}]{Paul Hayne}
is an Associate Professor in the Department of Astrophysical and Planetary Sciences, and the Laboratory for Atmospheric and Space Physics (LASP) at the University of Colorado Boulder. He earned his B.S. and M.S. in Geophysics from Stanford University, and his Ph.D. in Geophysics and Space Physics from UCLA. At LASP, he directs the Exploration of Planetary Ices and Climates (EPIC) group, which researches interactions between the surfaces and atmospheres of icy planets and moons throughout the solar system using data from deep space missions. 
\end{IEEEbiography}

\begin{IEEEbiography}
[{\includegraphics[width=1in,height=1.25in,clip,keepaspectratio]{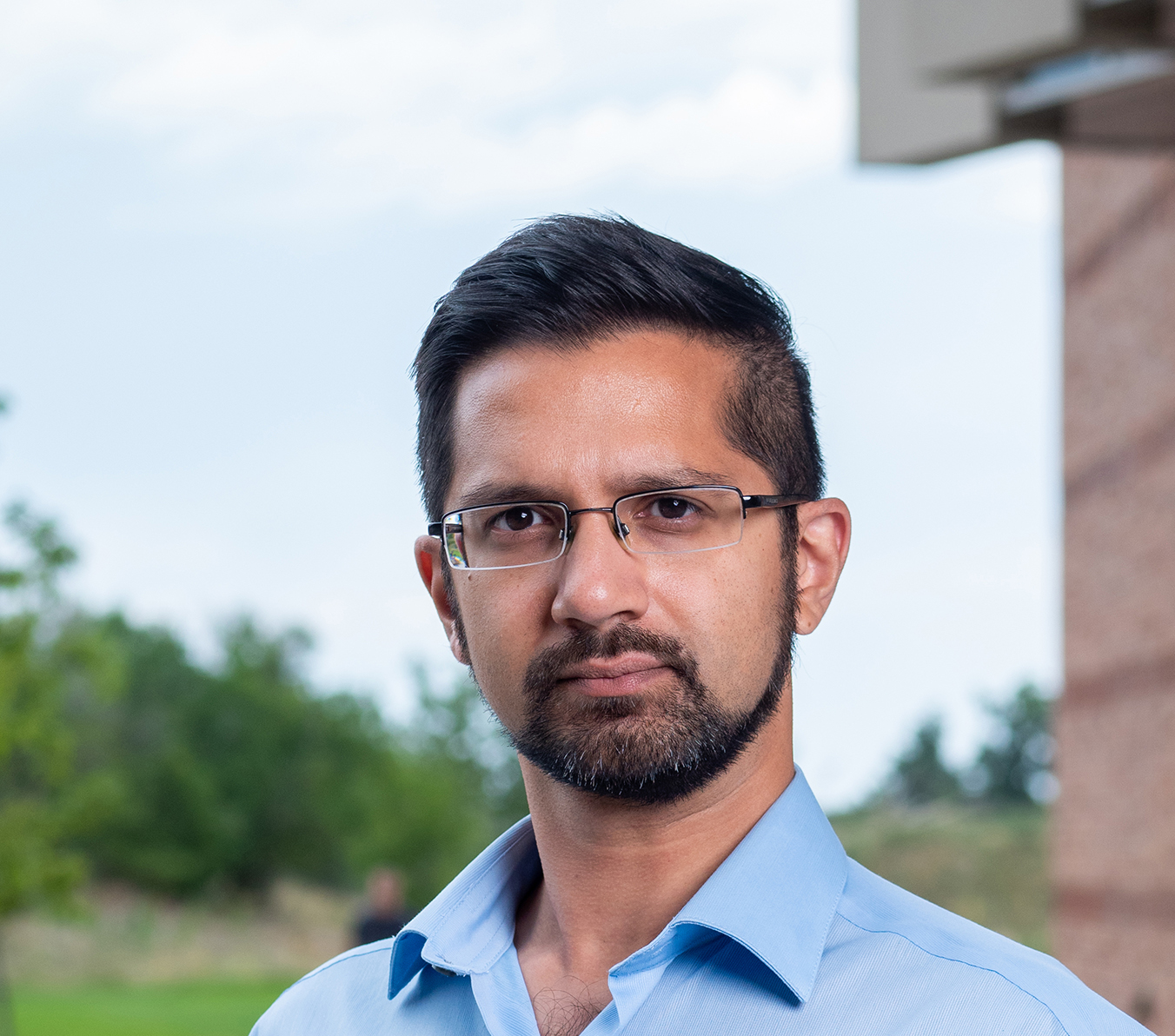}}]{Nisar Ahmed}
is an Associate Professor and H.J. Smead Faculty Fellow in the Smead Aerospace Engineering Sciences Department at the University of Colorado Boulder. 
He earned his B.S. in Engineering from Cooper Union in New York City in 2006, and his Ph.D. in Mechanical Engineering from Cornell University in Ithaca, NY in 2012. 
He 
directs the Cooperative Human-Robot Intelligence (COHRINT) Lab, which researches probabilistic modeling, estimation and control of autonomous systems, human-robot/machine interaction, sensor fusion, and decision-making under uncertainty. 
\end{IEEEbiography}

\vspace{11pt}

\vfill

\end{document}